# CSCF: A Chaotic sine cosine firefly Algorithm for practical application problems

Bryar A. Hassan[1,2]

[1]Kurdistan Institution for Strategic Studies and Scientific Research, Sulaimani, Iraq.

[2]Department of Computer Networks, Technical College of Informatics,
Sulaimani Polytechnic University, Sulaimani, Iraq.

Email: bryar.hassan@kissr.edu.krd

## Abstract

Recently, numerous meta-heuristic based approaches are deliberated to reduce the computational complexities of several existing approaches that include tricky derivations, very large memory space requirement, initial value sensitivity etc. However, several optimization algorithms namely firefly algorithm, sine cosine algorithm, particle swarm optimization algorithm have few drawbacks such as computational complexity, convergence speed etc. So to overcome such shortcomings, this paper aims in developing a novel Chaotic Sine Cosine Firefly (CSCF) algorithm with numerous variants to solve optimization problems. Here, the chaotic form of two algorithms namely the sine cosine algorithm (SCA) and the Firefly (FF) algorithms are integrated to improve the convergence speed and efficiency thus minimizing several complexity issues. Moreover, the proposed CSCF approach is operated under various chaotic phases and the optimal chaotic variants containing the best chaotic mapping is selected. Then numerous chaotic benchmark functions are utilized to examine the system performance of the CSCF algorithm. Finally, the simulation results for the problems based on engineering design are demonstrated to prove the efficiency, robustness and effectiveness of the proposed algorithm.

## 1. Introduction

In recent decades, numerous algorithms have proposed to overcome various optimization problems in the field of engineering [1]. These optimization problems determine the value of a few parameters under specific circumstances for optimizing the objective function. In general, the objective function is a specific characteristic that provides a minimal or maximal solution based on the problem. Therefore, to attain best optimistic solutions, the optimizations are broadly utilized in various applications such as industrial design, manufacture design, design analysis and engineering design etc. [2]. There occur several optimization problems while obtaining optimal solution and these optimization problems are categorized into several types namely dynamic or static, continuous or discrete, single-objective or multi-objective as well as constrained or unconstrained. Hence, to enhance the accuracy and the efficiency of such optimization problems, several research scholars depend upon meta-heuristic algorithms for easy implementation, gradient information and to avoid or bypass local optimization problem [3].

At the same time, the meta-heuristic algorithm plays a significant role in engineering field due to its wide range of challenges. But the nature-inspired meta-heuristic algorithms are relatively straight forward and are inspired mostly by unsophisticated ideas. Normally, the meta-heuristic algorithms are divided into four main categories namely swarm-based algorithm, evolutionary-based algorithm; human behaviour based algorithm as well as physic-based algorithm [4]. Optimization algorithms such as Differential Evolution (DE) algorithm [5], Evolution Strategy (ES) algorithm [6], backtracking search optimization Algorithm (BSA) [44][45], Genetic Algorithm (GA) [7] etc. are categorized under evolutionary-based technique [36] [38] [39]. Few algorithms namely FireFly (FF) algorithm [8] [34] [35], Particle Swarm Optimization (PSO) algorithm [9] [37], Artificial Bee Colony (ABC) algorithm [10] are categorized under swarm intelligence approach; Gravitational Search (GS) algorithm [11], Black Hole (BH) algorithm [12] are few algorithms that are characterized under physic-based techniques. The human-based behaviour comprises of Mine Blast (MB) algorithm [13], League Championship (LC) algorithm [14] etc. The evolutionary algorithm imitates the behaviour of evolutionary processes thereby producing a global optimal offspring value. Then the candidate solutions are enhanced and the iterative processes are continued until it satisfies the terminating criteria [15] [30].

In the present study, a novel Chaotic Sine Cosine Firefly (CSCF) algorithm is developed with numerous variants to solve optimization problems. Then, the chaotic form of two algorithms namely the sine cosine algorithm (SCA) and the Firefly (FF) algorithms are integrated to improve the convergence speed and efficiency thereby minimizing several complexity issues. In addition to this, the proposed CSCF approach is operated under various chaotic phases and the optimal chaotic variants containing the best chaotic mapping is selected to determine the efficiency of the system.

This research study fulfils four main objectives

- Demonstrating a novel Chaotic Sine Cosine Firefly (CSCF) algorithm to improve the convergence speed and efficiency thereby minimizing several complexity issues
- Developing numerous variants such as Variant- I, Variant- II, Variant- III, Variant- IV and Variant- V of the novel CSCF algorithm for chaotic tuning of numerous parameters.
- Utilizing numerous chaotic benchmark functions to examine the system performance of the CSCF algorithm.
- Demonstrating engineering design problems to prove the efficiency, robustness and effectiveness of the CSCF algorithm.

The rest of the paper is structured as follows. Section 2 provides the basics of two different types of algorithms namely firefly (FF) algorithm and Sine cosine algorithm (SCA). Section 3 describes the proposed CSCF algorithm by forming diverse variants to tune the parameters. Then section 4 demonstrates the chaotic and benchmark functions to determine the efficiency of the system; the comparative analysis of various approaches with the proposed CSCF algorithm and finally the engineering design problems are solved. Section 5 concludes the article.

## 2. Related Literal Works

### *2.1 Improved Firefly (IFF) Algorithm*

- *Standard Firefly Algorithm*

The Firefly algorithm was first developed by Xin-She at the University of Cambridge. The firefly algorithm imitates the characteristics of the firefly as well as its locomotion activities [16]. This firefly algorithm is considered to be a more efficient approach in finding solutions for various crucial engineering related issues because of very high exploration capability, its brightness and flashlight capability. Also, the firefly algorithm works under the principle of bionics [40]. The most effective firefly algorithm is selected to obtain the best optimal value despite the complex and non-linear design. In general, each individual firefly has the capability to flash its light to attract the adjacent firefly thereby providing arbitrary solutions.

A. *Purpose for flashing light*

- Attracting the partner for mating since every firefly is unisexual in nature

- The attracting capability is proportional to the brightness. The firefly also utilizes its flashing light capability to attract the prey for survival.
- Moreover, the firefly uses its flashing light to protect themselves from other enemies [17].

*B. Light Intensity variation and Attraction Capability*

The light intensity variation and the attraction capability play a significant role in the firefly algorithm. The fitness value is determined by the light intensity. It also has the capability to deal with several multi optimization problems and are highly non-linear [41]. Here, the firefly with high or low intensity gets attracted with the neighbouring firefly having high or low intensity [18]. Let us consider, $D_{XY}$ be the distance among two fireflies namely $X$ *and* $Y$. Furthermore, the intensity of the light diminishes concerning the distance from the source; also the media absorbs the light. Then as per the law of square inverse, the expression for the intensity of the light is represented in equation (1).

$$I(D) = \frac{I_S}{D_{XY}^2} \tag{1}$$

From equation (1), the source intensity is represented as $I_S$. Then the expression for the intensity of the light $I_L$ that varies concerning the distance $D_{XY}$ is mentioned in equation (2). Therefore,

$$I_L = I_0 e^{-\beta D_{XY}} \tag{2}$$

From the above equation, $\beta$ is the absorption coefficient of the fixed light. The initial intensity of the light is denoted as $I_0$.

Each firefly contains a very strong attractive capability. This implies the strong attracting behaviour of a firefly over neighbouring firefly groups. Based on the distance between the two fireflies namely $X$ *and* $Y$, the attractive capability is varied. We know that the attractive capability of the firefly is directly proportional to the intensity of the light of the neighbouring fireflies. Therefore, the expression based on the attractive function is delineated in the following equation.

$$\alpha = \alpha_0 e^{-\beta D_{XY}^2} \tag{3}$$

From equation (3), the attractive capability at distance $D_{XY} = 0$ is denoted as $\alpha_0$. For a fixed, characteristic length becomes

$$\Psi = \beta - 1/M \to 1, M \to \infty \tag{4}$$

The Cartesian distances among two different fireflies $X\ and\ Y$ are denoted as $P_X\ and\ P_Y$ correspondingly. Therefore, the Cartesian distance formula for two fireflies is expressed in the following expression.

$$D_{X,Y} = \|P_X - P_Y\| = \sqrt{\sum_{R=1}^{\alpha}(P_X^R - P_Y^R)^2} \tag{5}$$

Then the movement of attraction from one fly to another fly i.e. $X\ and\ Y$ is characterised in equation (6). The determination of firefly movement with respect to the attracting capability is defined as,

$$P_X^{R+1} = P_X^R + \alpha_0\, e^{-\beta D_{XY}^2} (P_Y - P_X) + J\eta \tag{6}$$

From equation (6), $\alpha_0 e^{-\beta D_{XY}^2} (P_Y - P_X)$ be the attractive term and $J\eta$ be the term containing random variable ranges from [0, 1].

- *Improved FireFly Algorithm*

In the firefly algorithm, the best optimal value provides the firefly with very high brightness. The model of other classical approaches suffers from the trapping of local minima due to the non-linear design strategy. Therefore Improved Firefly algorithm was developed to improve the trapping of local minimum value. Here, an additional term is added to the standard firefly algorithm to achieve better randomness and efficiency of the firefly. Then the difference among the arbitrary firefly and the $X\ th$ position of the firefly is obtained thereby achieving the effective randomness of the firefly [19].

The brightest firefly found among the firefly group is said to be known as Best firefly $B_F$. The random number is denoted as $R_4$ and the value ranges from [0, 1]. Even though the computational complexity is high in case of improved firefly optimization, the local trapping quality is good for tuning the $J$ and $K$ parameter value. Therefore, the modified firefly algorithm is referred to as improved firefly algorithm. Then the updated expression is determined in the following section.

$$P_X^{R+1} = P_X^R + \alpha_0 e^{-\beta D_{XY}^2} (P_Y - P_X) + J\eta + K(P_A - P_X) \tag{7}$$

Where, $A \neq X \; and \; Y$. $K(P_A - P_X)$ be the term containing random variable ranges from [0, 1]. The pseudo-code for improved firefly algorithm is delineated as follows.

***Pseudo-code 1:* Improved Firefly Algorithm**

**Input:** Size of the population

**Output:** Best optimal solution

Parameter setting of firefly algorithm;
Generation of initial population;
Determination of light intensity;
**while** $(T \leq Max \;\; Iteration)$
Define absorption coefficient $\beta$;
**for** $X = 1{:}n$ (n fireflies)
    **for** $Y = 1{:}n$ (n fireflies)
        **if** $I_Y > I_X$
            move firefly towards $Y$;
        **end if**
            Vary attractiveness with distance $D_{XY}$ via $e^{-\beta D_{XY}^2}$;
            Evaluation of new solutions and updating of light intensity;
        **end for**
    **end for**
Firefly ranking and determination of best solution;
$T = T + 1$
**end while**

- *Chaotic Firefly algorithm*

The fireflies are also referred to as lighting bugs that are found during the night time particularly in the summer season [31]. In chaotic optimization approaches, the chaotic firefly variables replace the random variables [32]. The chaotic firefly algorithm selects the initial population of the search algorithm. The absorption coefficient $\beta$ found in the solution space and the firefly positions is updated by employing the chaotic sequence demonstrated by the chaotic maps that are represented in Table 1. From equation (6), the step size $J$ affects the random vector $\eta$ and the chaotic time series replaces the third term and the mathematical expression is obtained as follows.

$$\eta_X = CHAOS_X^K \tag{8}$$

From equation (8), the chaotic maps are represented by $CHAOS_X^K$; where the superscript $K$ represents the type of chaotic map to be determined. In a similar way, the attractive term $\alpha_0$ from equation (6) is substituted by the chaotic term that is represented in the following equation.

$$\alpha_0 = \alpha_0 CHAOS_X^K \tag{9}$$

The random motion of the firefly plays a significant role in determining the candidate [where $\alpha_0$(attractive term) relies on $\beta$ (light absorption coefficient)] from the population. In addition to these two limiting cases $(\beta \to 0;\ \alpha \to \alpha_0)$ are formulated while determining $\beta$. Therefore, the entire fireflies can spot one another and they start moving randomly when $\beta$ becomes $\infty$.

$$P_X^{R+1}(T+1) = P_X^R(T) - \alpha_0\, e^{-\beta D_{XY}^2} P_X(T) \tag{10}$$

The light absorption coefficient then employs in characterizing the dissimilarities in the attractiveness as well as the values are vitally imperative to determine the convergence capability and the speed of the firefly algorithm.

| ***Pseudo-code 2:*** **Sine Cosine Algorithm** |
|---|
| **Input:** Population size<br>**Output:** Best solution |
| 1: Initializing the set of solution or search agents;<br>2: **do**<br>3: Evaluation of search agents or solution by employing the fitness function;<br>4: Updation of so far obtained best solution;<br>5: Updation of $R_1, R_2, R_3$ *and* $R_4$ ;<br>6: Position updation using equation (13);<br>7: **while** $(T \leq Max\ \ Iteration)$ **do**<br>8: **Return** |

***2.2 Sine Cosine (SC) Algorithm***

In general, the optimization approach based on population initiates its optimization method containing a random solution. The Sine-Cosine algorithm was first developed by Mirjalili in the year of 2016 for solving several optimization issues [20]. The SC algorithm utilizes sine and cosine functions to determine the best optimal solution. In SCA, the distance and the movement among each feature solution and the best member are affected. Therefore, the SCA employs a balance equation utilizing two phases namely the exploration phase and the exploitation phase. The solutions are changed randomly in the exploration case whereas, in the exploitation phase, the random variables are less.

The updating equation for both the exploitation and the exploration phase is expressed in the following equation.

$$P_X^{T+1} = P_X^T + R_1 \sin(R_2) \left| R_3 Z_X^T - P_X^T \right| \tag{11}$$

$$P_X^{T+1} = P_X^T + R_1 \cos(R_2) \left| R_3 Z_X^T - P_X^T \right| \tag{12}$$

From equation (11) and (12), $P_X^T$ signifies the position of the current solution; where $X$ represents the dimension and $T$ represents the iteration. The random numbers are represented as $R_1, R_2$ *and* $R_3$. The position of the destination point in *Xth* dimension is represented as $Z_X$ ; $|\ |$ represents the absolute value.

Then the combined equation based on sine cosine algorithm is represented as follows. Therefore,

$$P_X^{T+1} = \begin{cases} P_X^T + R_1 \sin(R_2) \left| R_3 Z_X^T - P_X^T \right|; \ R_4 < 0.5 \\ P_X^T + R_1 \cos(R_2) \left| R_3 Z_X^T - P_X^T \right|; R_4 \geq 0.5 \end{cases} \tag{13}$$

From equation (13), the random number is denoted by $R_4$ and the random values may range from [0, 1]. From equation (11)-(13), the region of next position that present in between the destination and the solution is denoted by $R_1$; The movement outwards the destination is represented by $R_2$ . The random weight is represented utilizing two different constraints determined in the following equation.

$$\begin{cases} if \ \ R_3 > 1; \quad stochastically \ \ emphazized \\ if \ \ R_3 < 1; \quad stochastically \ \ deemphazized \end{cases} \tag{14}$$

Moreover, the exploration and the exploitation phase f the algorithm is to be balanced for finding the search space region. Therefore, equation (11) to (12) is modified for balancing both the phases that are determined in the following equation. Therefore,

$$R_1 = A - T\frac{A}{t} \tag{15}$$

From equation (15), the maximum iteration number and the current iterations are represented by $t$ and $T$ respectively and the constant term is denoted as $A$.

The general procedure of the sine cosine algorithm is represented in the following section.

- *Chaotic Sine Cosine Algorithm*

In this section, the parameters $R_1$, $R_2$ and $R_3$ of equation (13) are modulated using chaotic maps during iterations that are described in section 3 [equations (18 to 20)]. The meta-heuristic algorithms use a conventional method to find the best optimal solutions that are based on iteration; also it relies on random solutions to replicate the naturally occurring phenomenon [42]. More clear that, there occurs a major shortcoming based on the solution outcome and the convergence speed since these solutions rely upon random parameters. Consequently, the random parameters are replaced with the chaotic parameters also; numerous chaotic mapping functions are employed to enhance the overall performances of the optimization approach [21]. In addition to this, new parameters are introduced to replace the random numbers and functions with various deterministic numbers [22]. Also, the standard distributive functions namely Gaussian distribution [33] and uniform distributions are replaced with the non-standard distributive functions namely chaotic based optimization algorithms. Moreover, the chaotic forms of the sine cosine algorithms are employed in boosting the performances of the sine cosine algorithm [43].

## 3. Proposed CSCF Algorithm

Numerous meta-heuristic based approaches are deliberated to eliminate the computational complexities of several existing approaches namely complex and tricky derivations, the requirement of very large memory space, initial value sensitivity etc. In general, the meta-heuristic based approaches are deliberated to reduce the computational complexities of several existing approaches that contain complex and tricky derivations, the requirement of very large memory space, initial value sensitivity etc. However, several optimization algorithms such as firefly algorithm, sine cosine algorithm, particle swarm optimization algorithm have few drawbacks such as computational complexity, convergence speed etc. So to overcome such shortcomings, this paper aims in developing a novel Chaotic Sine Cosine Firefly (CSCF) algorithm with numerous variants to solve optimization problems.

In this paper, the improved firefly algorithm and the sine cosine optimization algorithms are integrated to form a Sine Cosine Firefly approach. Then the integrated algorithm aims in hybridizing the chaotic algorithm [i.e. Chaotic Sine Cosine Firefly (CSCF) approach] containing various chaotic mapping functions. The proposed CSCF approach is operated under various chaotic phases and the optimal chaotic variants containing the best chaotic mapping is selected. The general architecture of the proposed CSCF algorithm is represented in fig.1. The initial step involves in parameter initialization followed by the random initialization of the function. Then the fitness function is evaluated and if the trial is less than the limit the chaotically tuned *Jth and Kth* of the firefly algorithm is formulated to obtain the best optimal solution; else the chaotically tuned $R_1$, $R_2$ *and* $R_3$ of the sine cosine algorithm is formulated to obtain the best optimal solution. The boxes that are highlighted characterize the new variants of the original algorithm. The diverse variants of CSCF approaches are delineated in the following section.

### *3.1 Variants of CSCF approach*

The following subsections describe numerous variants of the CSCF (sine cosine algorithm and firefly algorithm) approach in accordance with the tuned parameters.

- *Variant- I*

The parameter $J$ of equation (7) is modified by chaotic maps (CM). Therefore, the new version of equation (7) is determined in the following equation.

$$P_X^{R+1} = P_X^R + \alpha_0 e^{-\beta D^2}(P_Y - P_X) + J^{CHAOS(.)}(\eta) + K(P_A - P_X) \tag{16}$$

From equation (16), the chaotic random movement of the firefly is denoted as $J^{CHAOS(.)}$ and are determined by $J^{CHAOS(.)} = J.CM_I$. Here, $J$ is a fixed value in standard firefly, while in variant-I, it evolves chaotically.

- *Variant- II*

In this version, the parameter $K$ of equation (7) is modified such that it is changed chaotically using the chaotic maps (CM). Therefore,

$$P_X^{R+1} = P_X^R + \alpha_0 e^{-\beta D^2}(P_Y - P_X) + J(\eta) + K^{CHAOS(.)}(P_A - P_X) \quad (17)$$

From the above equation, $K^{CHAOS(.)}$ represents the chaotic fractional difference between the arbitrary fireflies. The standard firefly optimization algorithm generates the random position; whereas the chaotic firefly generates according to the chaotic maps.

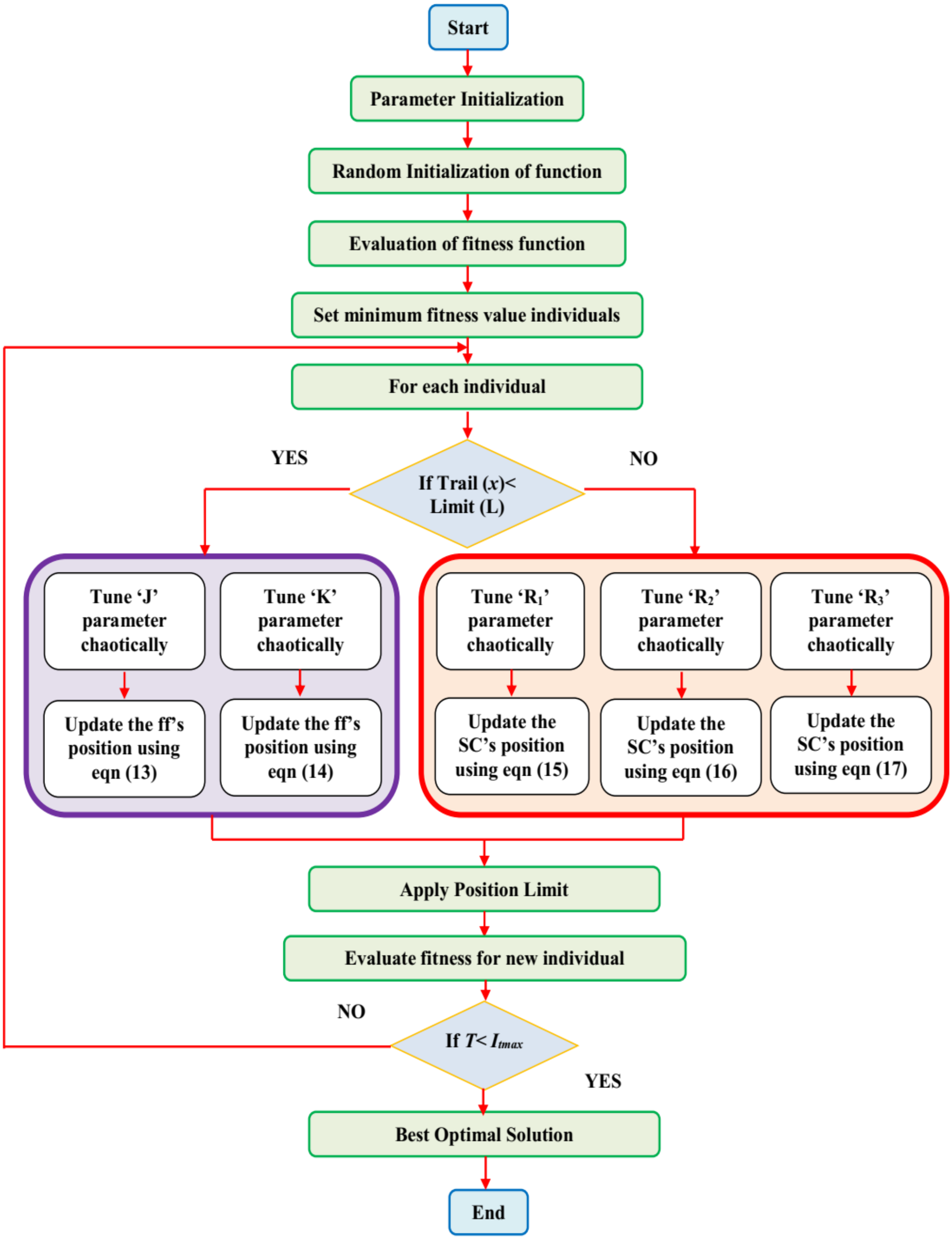

**Fig.1.** Flow chart representation of a CSCF algorithm

- *Variant- III*

The parameter $R_1$ of equation (13) is modulated using chaotic maps during iterations. Thus equation (13) is changed to,

$$P_X^{T+1} = \begin{cases} P_X^T + R_1^{CHAOS(.)} \sin(R_2) \left| R_3 Z_X^T - P_X^T \right|; \; R_4 < 0.5 \\ P_X^T + R_1^{CHAOS(.)} \cos(R_2) \left| R_3 Z_X^T - P_X^T \right|; R_4 \geq 0.5 \end{cases} \quad (18)$$

From equation (18), in standard sine cosine algorithm, $R_1$ is created randomly between 0 and 1; while in variant III, it is a chaotic number between 0 and 1.

- *Variant- IV*

In this version, $R_2$ the parameter is modified such that it is chaotically altered using CM. Then,

$$P_X^{T+1} = \begin{cases} P_X^T + R_1 \sin(R_2^{CHAOS(.)}) \left| R_3 Z_X^T - P_X^T \right|; \; R_4 < 0.5 \\ P_X^T + R_1 \cos(R_2^{CHAOS(.)}) \left| R_3 Z_X^T - P_X^T \right|; R_4 \geq 0.5 \end{cases} \quad (19)$$

Where, in the standard SC algorithm, the random number varies from 0 to 1. While in variant IV, it is a chaotic number between 0 and 1.

- *Variant- V*

The parameter $R_3$ of equation (13) is modulated using chaotic maps during iterations. Thus equation (13) becomes,

$$P_X^{T+1} = \begin{cases} P_X^T + R_1 \sin(R_2) \left| R_3^{CHAOS(.)} Z_X^T - P_X^T \right|; \; R_4 < 0.5 \\ P_X^T + R_1 \cos(R_2) \left| R_3^{CHAOS(.)} Z_X^T - P_X^T \right|; R_4 \geq 0.5 \end{cases} \quad (20)$$

From the above equation, $R_3^{CHAOS(.)}$ is the chaotic random value that ranges from 0 to 1.

**4. Result and Discussions**

In this section, various experiments are conducted to evaluate the efficiency and the performances of the Chaotic Sine Cosine Firefly (CSCF) algorithm. Owing to its hypothetical nature, various chaotic functions and benchmark functions are discussed to obtain better optimal results. In addition to this, the proposed CSCF algorithms are compared with several other optimization algorithms such as FireFly (FF) algorithm [19], Sine Cosine Algorithm (SCA) [20], Particle Swarm Optimization (PSO) approach [23], Artificial Bee Colony (ABC) optimization algorithms [24] to evaluate the effectiveness of the CSCF algorithm. Furthermore, followed by the comparison of optimization algorithms, a detailed description of the real-time engineering applications are delineated in the following section. The experimental analysis is carried out under the platform of MATLAB R2016a by using the operating system as Windows 10. The simulations are done on a central processing unit containing Intel Core (TM) i7-6700HQ CPU @ 2.60 GHz with 8G of memory. Then the chaotic benchmark functions and the test functions are explained as follows.

**Table 1.** Chaotic Mapping (CM) description and its functions

| No | Name of the Maps | Mapping Functions |
|---|---|---|
| Chaotic Map 1 | Logistic | $Z_{J+1} = AZ_J(1-Z_J)$ |
| Chaotic Map 2 | Tent | $Z_{J+1} = \begin{cases} \beta - Z_J; & Z_J < 0.5 \\ \beta(1-Z_J); & 0.5 \le Z_J \end{cases}$ |
| Chaotic Map 3 | Sinusoidal | $Z_{J+1} = AZ_J^2 \sin(\pi Z_J)$ |
| Chaotic Map 4 | Gauss | $Z_{J+1} = \begin{cases} 0; & Z_J = 0 \\ \frac{1}{Z_J} Mod(1); & Z_J \in [0,1] \end{cases}$ Where, $\frac{1}{Z_J} Mod(1) = \frac{1}{Z_J} - \left[\frac{1}{Z_J}\right]$ |
| Chaotic Map 5 | Circle | $Z_{J+1} = Z_J + B - \left(\frac{A}{2\pi}\right)\sin(2\pi J)\mathrm{mod}(1);$ $where\ A = 0.5\ and\ B = 0.2$ |
| Chaotic Map 6 | Sinus | $Z_{J+1} = 2.3(Z_J)^2 \sin(\pi Z_J)$ |
| Chaotic Map 7 | Iterative | $Z_{J+1} = \sin\left(\frac{A\pi}{Z_J}\right); \quad A \in [0,1]$ |
| Chaotic Map 8 | Chebyshev | $Z_{J+1} = 2.3(Z_J)^2 \sin(\pi Z_J)$ |
| Chaotic Map 9 | Henon | $Z_{J+1} = 1 - P(Z_J)^2 + QZ_J - 1$ |
| Chaotic Map 10 | Intermittency | $Z_{J+1} = \begin{cases} \delta + Z_J + BZ_J^N; & 0 < Z_J \le P \\ \frac{Z_J - P}{1-P}; & P < Z_J \le 1 \end{cases}$ |
| Chaotic Map 11 | Singer | $Z_{J+1} = \alpha(7.8\ Z_J - 23.3\ Z_J^2 + 28.7\ Z_J^3 - 13.3\ Z_J^4)$ |
| Chaotic Map 12 | Sine | $Z_{J+1} = \frac{A}{4}\sin(\pi Z_J); \quad 0 < A \le 4$ |

**Table 2.** Benchmark functions

| Fn | Mathematical Equations | $D_M$ | Ranges | $F_{OV}$ |
|---|---|---|---|---|
| $Fn^1$ | $Fn^1 = -20Exp(-0.2\sqrt{\frac{1}{J}\sum_{K=1}^{J} P_M^2}) -$ $Exp(\frac{1}{J}\sum_{K=1}^{J} COS(2\pi P_I) + 20 + E$ | 20 | $[-30,30]$ | 0 |
| $Fn^2$ | $Fn^2 = \frac{1}{4000}\sum_{K=1}^{J} P_K^2 - \prod_{K=1}^{J} \cos(\frac{P_1}{\sqrt{K}}) + 1$ | 20 | $[-600,600]$ | 0 |
| $Fn^3$ | $Fn^3 = 30 + \sum_{K=1}^{J} \lfloor P_K \rfloor$ | 20 | $[-5.12,5.12]$ | -6J+30 |
| $Fn^4$ | $Fn^4 = \sum_{K=1}^{J} \sin(10 Log(P_K))$ | 20 | [0.25,10] | -J |
| $Fn^5$ | $Fn^5 = \sum_{K=1}^{J} \left\| P_K^5 - 3P_K^4 + 4P_K^3 + 2P_K^2 - 10P_K - 4 \right\|$ | 20 | $[-10,10]$ | 0 |
| $Fn^6$ | $Fn^6 = P_K^2$ | 20 | [-100,100] | 0 |
| $Fn^7$ | $Fn^7 = \sum_{K=1}^{J} \left( \sum_{K-1}^{K} P_K \right)^2$ | 20 | [-100,100] | 0 |
| $Fn^8$ | $Fn^8 = \sum_{K=1}^{J} \left( \left\| P_K + 0.5 \right\| \right)^2$ | 20 | [-10,10] | -3.214 |
| $Fn^9$ | $Fn^9 = \sum_{K=1}^{J} - P_K \sin\left( \sqrt{\left\| P_K \right\|} \right)$ | 20 | [-5.12, 5.12] | 0 |
| $Fn^{10}$ | $Fn^{10} = \sum_{K=1}^{J} \left[ \left( P_K^2 - 10\cos(2\pi P_K) \right) + 10 \right]$ | 20 | [-200,200] | 0 |
| $Fn^{11}$ | $Fn^{11} = 4P_1^2 - 2.1P_1^4 + \frac{1}{3}P_1^6 + P_1 P_2$ $-4P_2^2 - 4P_2^4$ | 20 | [-5,5] | -1.6428 |
| $Fn^{12}$ | $Fn^{12} = [1 + (P_1 + P_2 + 1)^2 (19 - 14P_1 + 3P_1^2$ $-14P_2 + 6P_1P_2 + 3P_2^2)]$ $*[30 + (2P_1 - 3P_2)^2 (18 - 32P_1$ $+12P_1^2 + 48P_2 - 36P_1P_2 + 27P_2^2)]$ | 20 | [-3,3] | 3 |
| $Fn^{13}$ | $Fn^{13} = -\sum_{K=1}^{10} \left[ (P - Y_I)(P - Z_K)^T + Z_K \right]^{-1}$ | 20 | [0,20] | -10.4673 |
| $Fn^{14}$ | $Fn^{14} = \sum_{K=1}^{J} KP_K^4 + Random[0,1)$ | 20 | [-1.28,1.28] | 0 |
| $Fn^{15}$ | $Fn^{15} = \sum_{K=1}^{J} - P_I \sin(\sqrt{P_K})$ | 20 | [-500,500] | 0 |
| $Fn^{16}$ | $Fn^{16} = Max\left\{ \left\| P_K \right\|, 1 \le K \le J \right\}$ | 20 | [-600,600] | 1 |
| $Fn^{17}$ | $Fn^{17} = \sum_{K=1}^{J=1} [100(P_{K=1} - P_K^2)^2 + (P_K - 1)^2]$ | 20 | [-65, 65] | -209 |
| $Fn^{18}$ | $Fn^{18} = \frac{1}{25}\sum_{K=1}^{6} P_I^2 - \prod_{K=1}^{6} \cos(\frac{P_1}{\sqrt{K}}) + 1$ | 20 | [0,1] | -3.33 |
| $Fn^{19}$ | $Fn^{19} = \frac{\pi}{J}\left\{10\sin(\pi P_1) + \sum_{I=1}^{N} \begin{matrix}(P_K - 1)^2[1+10\sin^2((\pi P_{K-1})]+ \\ (P_J - 1)^2 + \sum_{K=1}^{N} U(P_K,10,100,4)\end{matrix}\right\}$ | 20 | [-50,50] | 0 |
| $Fn^{20}$ | $Fn^{20} = 0.1\left\{\begin{matrix}\sin^2(3\pi P_1) + \sum_{K=1}^{J}(P_I - 1)^2[1+ \\ \sin^2(3\pi P_1 + 1) \\ +(P_J - 1)^2[1+\sin^2(2\pi P_J)]+ \\ \sum_{K=1}^{J} U(P_K,5,100,4)\end{matrix}\right\}$ | 20 | [-50,50] | 0 |

**Table 3.** Comparative analysis of CSCF with SCF algorithm

| **Test fn.** | **CSCF** | | | | **SCF** | | | |
|---|---|---|---|---|---|---|---|---|
| | **μ** | **σ** | **B** | **W** | **μ** | **σ** | **B** | **W** |
| $Fn^{1}$ | 1.23E+02 | 3.27E+03 | **5.27 E+03** | **6.37 E+02** | 7.37 E+02 | **2.17 E+05** | 6.38 E+02 | **9.52E+02** |
| $Fn^{2}$ | **2.46 E+03** | **2.49E+04** | 2.38 E+05 | **3.49 E+02** | **5.27 E+01** | 3.49 E+06 | **5.30 E+02** | 7.39 E+05 |
| $Fn^{3}$ | 2.62 E+03 | **6.38E+05** | **1.56 E+06** | 4.58 E+03 | 4.32 E+03 | 1.23 E+06 | 2.47 E+02 | 2.37 E+06 |
| $Fn^{4}$ | **1.47 E+04** | 4.50E+03 | **2.59 E+07** | **1.38 E+04** | 2.48 E+04 | **2.48 E+06** | 2.58 E+03 | **5.33 E+02** |
| $Fn^{5}$ | **3.65 E+02** | **3.59E+03** | 1.27 E+06 | 2.50 E+04 | **4.48 E+03** | 3.47 E+07 | 4.58 E+04 | 2.45 E+03 |
| $Fn^{6}$ | 2.47 E+02 | **1.49 E+04** | **2.50 E+04** | **5.37 E+05** | 6.58 E+05 | 4.45 E+08 | **5.68 E+05** | 4.57 E+04 |
| $Fn^{7}$ | **1.69 E+02** | 5.49 E+03 | 3.67 E+03 | **4.39 E+06** | 8.37 E+04 | 2.24 E+03 | 2.49 E+06 | **3.27 E+06** |
| $Fn^{8}$ | 1.50 E+02 | **1.37 E+03** | **4.28 E+03** | 2.54 E+07 | **3.28 E+04** | **1.34 E+02** | 7.48 E+04 | 1.36 E+07 |
| $Fn^{9}$ | **2.47 E+02** | 2.59 E+02 | 7.38 E+07 | **2.11 E+08** | 2.49 E+05 | 3.46 E+03 | 5.39 E+04 | 4.56 E+08 |
| $Fn^{10}$ | 3.46 E+02 | **5.28 E+02** | **3.31 E+02** | **4.87 E+02** | 1.25 E+02 | 4.61 E+02 | **3.18 E+02** | 6.27 E+02 |
| $Fn^{11}$ | 1.56 E+02 | **4.68 E+02** | 6.38 E+08 | 5.82 E+08 | 1.39 E+06 | 2.48 E+02 | 2.12 E+04 | **7.38 E+07** |
| $Fn^{12}$ | 2.15 E+04 | 7.47 E+06 | **2.46 E+09** | 7.95 E+23 | 2.68 E+07 | **4.57 E+03** | 3.34 E+07 | 3.27 E+06 |
| $Fn^{13}$ | **1.48 E+03** | 3.48 E+05 | **1.46 E+09** | **2.59 E+02** | **1.38 E+07** | 6.46 E+02 | 5.23 E+05 | 6.38 E+07 |
| $Fn^{14}$ | **2.48 E+05** | **1.37 E+07** | 2.27 E+07 | 1.22 E+11 | 2.59 E+07 | 4.68 E+02 | 2.37 E+07 | 3.27 E+09 |
| $Fn^{15}$ | 1.59 E+06 | 4.56 E+08 | **1.49 E+05** | **3.54 E+10** | 3.78 E+05 | **6.39 E+03** | 1.36 E+06 | **2.49 E+08** |
| $Fn^{16}$ | **1.67 E+05** | **3.28 E+07** | 3.48 E+05 | **4.28 E+08** | 2.40 E+05 | 7.35 E+05 | 1.48 E+04 | 1.27 E+05 |
| $Fn^{17}$ | **2.59 E+05** | **2.47 E+10** | **2.83 E+04** | 2.19 E+07 | **4.39 E+03** | 2.13 E+05 | **1.25 E+02** | 3.29 E+04 |
| $Fn^{18}$ | **3.86 E+07** | 1.39 E+09 | **2.49 E+07** | 1.37 E+04 | 2.59 E+03 | **1.46 E+03** | 2.49 E+02 | 2.19 E+03 |
| $Fn^{19}$ | 2.69 E+05 | **5.39 E+02** | 1.48 E+06 | **2.28 E+03** | 3.50 E+02 | 1.58 E+02 | 3.59 E+05 | **2.40 E+07** |
| $Fn^{20}$ | **1.78 E+05** | 3.28 E+03 | **5.30 E+05** | 1.38 E+02 | **2.28 E+06** | 1.30 E+02 | **4.37 E+07** | 1.38 E+05 |

**Table 4.** Comparative analysis of CSCF with SCF for various dimensions

| Test fn. | D=20 | | D=50 | | D=100 | |
|---|---|---|---|---|---|---|
| | **CSCF** | **SCF** | **CSCF** | **SCF** | **CSCF** | **SCF** |
| Fn[1] | 7.48 E+04 | 2.59 E+05 | 2.47 E+10 | 2.46 E+09 | 3.49 E+06 | **5.30 E+02** |
| Fn[2] | **5.27 E+03** | **2.74 E+02** | 5.82 E+08 | 1.39 E+06 | 2.48 E+02 | 2.58 E+03 |
| Fn[3] | 4.58 E+04 | 1.22 E+11 | **3.48 E+04** | **3.34 E+07** | 7.47 E+06 | 2.46 E+09 |
| Fn[4] | **4.57 E+03** | **3.54 E+02** | 2.59 E+07 | 5.39 E+04 | **3.48 E+05** | 1.46 E+09 |
| Fn[5] | 5.23 E+05 | 3.48 E+05 | **1.48 E+04** | 1.38 E+02 | 1.37 E+07 | 2.27 E+07 |
| Fn[6] | 7.38 E+07 | **1.37 E+02** | 1.25 E+02 | 6.38E+05 | **1.56 E+06** | 4.58 E+03 |
| Fn[7] | **4.67E+03** | 2.47 E+10 | **2.83 E+04** | **2.19 E+07** | 5.39 E+04 | 4.56 E+08 |
| Fn[8] | 3.48 E+05 | **6.46 E+02** | **5.39 E+04** | **6.38 E+07** | **7.47 E+06** | 2.46 E+09 |
| Fn[9] | **2.54E+03** | **4.68 E+02** | **2.12 E+04** | 3.27 E+09 | **3.48 E+05** | 1.46 E+09 |
| Fn[10] | 5.28 E+02 | 5.82 E+08 | 4.48 E+03 | 3.47 E+07 | 4.58 E+04 | 2.49 E+02 |
| Fn[11] | **2.49 E+03** | 1.37 E+04 | 2.59 E+03 | 2.49 E+08 | 1.37 E+07 | 2.27 E+07 |
| Fn[12] | **1.59 E+03** | **4.56 E+02** | **1.46 E+04** | 5.39 E+02 | **7.38 E+05** | 2.59 E+05 |
| Fn[13] | 1.67 E+05 | 3.28 E+07 | 2.27 E+07 | **2.46 E+07** | 7.95 E+23 | 2.49 E+08 |
| Fn[14] | **2.62 E+03** | **6.38E+02** | **4.28 E+04** | 2.40 E+05 | **7.35 E+05** | 1.48 E+04 |
| Fn[15] | **8.37 E+03** | 2.24 E+03 | **2.83 E+04** | 7.37 E+02 | 2.49 E+02 | 7.35 E+05 |
| Fn[16] | **3.28 E+03** | **1.34 E+02** | 7.39 E+05 | 5.27 E+01 | **3.59 E+05** | 2.13 E+05 |
| Fn[17] | 4.28 E+08 | **2.47 E+02** | 2.37 E+06 | 4.32 E+03 | **2.13 E+05** | **1.25 E+02** |
| Fn[18] | **2.19 E+03** | 2.58 E+03 | **5.33 E+04** | **6.38 E+07** | 1.46 E+03 | **2.49 E+02** |
| Fn[19] | **3.86 E+03** | **1.39 E+02** | **2.49 E+04** | 3.27 E+09 | **1.58 E+05** | 3.59 E+05 |
| Fn[20] | **7.95 E+03** | 5.49 E+03 | **3.67 E+04** | 4.39 E+06 | 7.38 E+07 | 2.11 E+08 |
| Total | 12 | 10 | 11 | 5 | 9 | 3 |

### *4.1 Chaotic Mapping (CM) and Benchmark functions*

In this section, numerous chaotic benchmark functions are utilized to improve the system performance of the CSCF algorithm [25]. The complex operations, its nature and several other possessions of these chaotic functions are obtained easily from the definitions. Then the chaotic benchmark functions for CSCF algorithm is obtained in Table 1. Table 2 represents the benchmark functions with three different models namely uni-modal test function, multi-modal test function and three fixed dimensions multi-modal test function. Here 20 benchmark functions are employed to investigate the system performances. In Table 2, *Fn* represents the test functions, $\mathbf{D_M}$ and $\mathbf{F_{OV}}$ signify the dimensional value and the optimal value respectively.

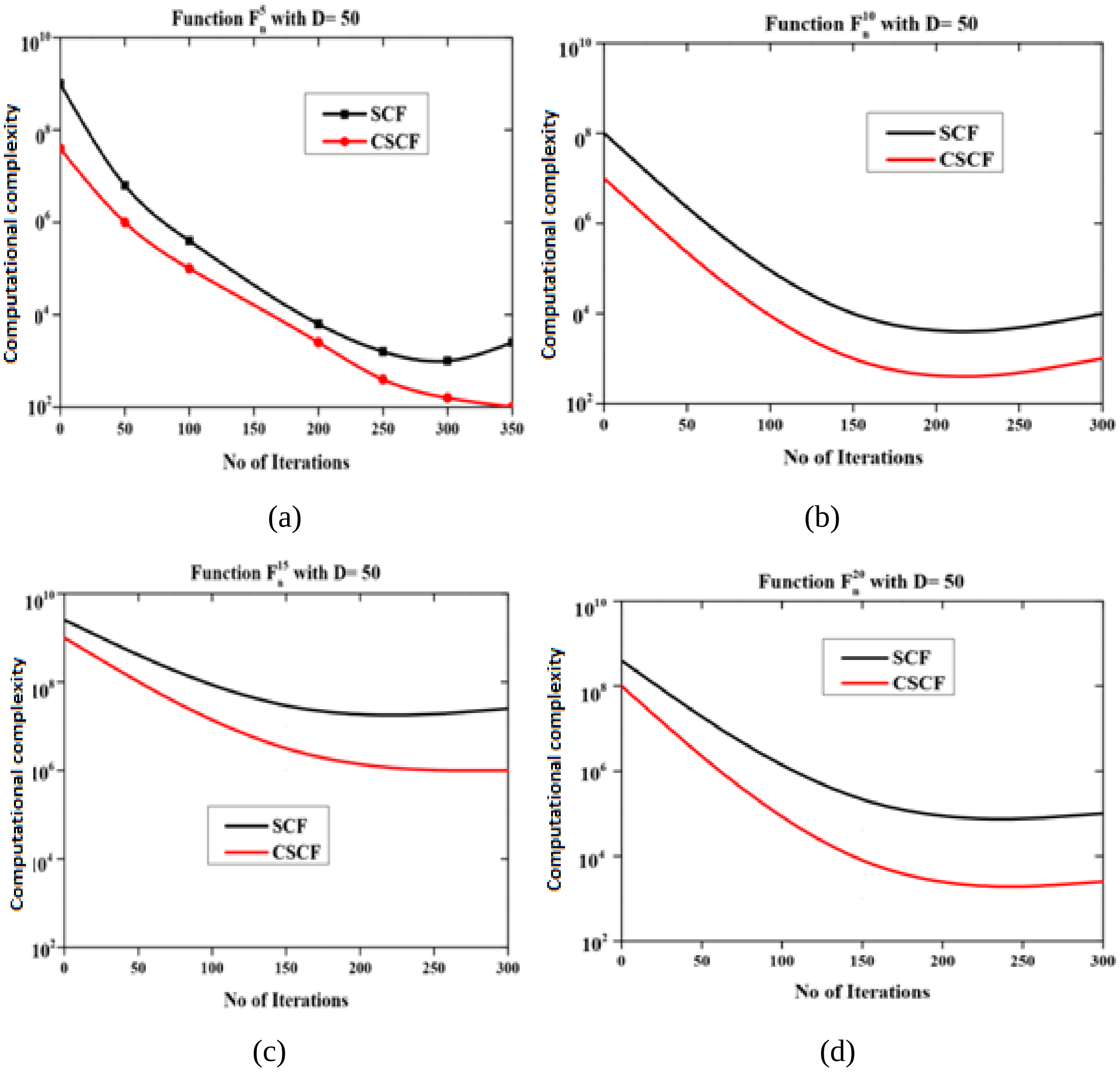

**Fig.2.** Graphical analysis of functions a) $Fn^5$, b) $Fn^{10}$, c) $Fn^{15}$and d)$Fn^{20}$ with D=50

### *4.2 Comparison of CSCF with Sine Cosine Firefly algorithm*

In this section, the chaotic sine cosine firefly is compared with sine cosine and firefly algorithms by using four different parameters namely mean *(μ)* standard deviation *(σ)*, best *(B)* and worst *(W)* values. For each functions, the size of the population is predetermined as 20 and the subsequent dimensional values may range from 20, 50 and 100 respectively. In addition to this, the maximum iteration value is set as 500. The comparative analyses of CSCF and sine cosine firefly algorithms are mentioned in Table 3. From the Table, the comparative analysis reveals that the proposed CSCF algorithm provides better performances when compared with SCF approach. Then Table 4 provides the comparative analysis of CSCF with SCF for different dimensions namely *D=20, D=50, D=100*. Then the convergence curve is compared for the proposed CSCF and SCF algorithms for various dimensions and the graphical analysis for four functions namely $Fn^5$, $Fn^{10}$, $Fn^{15}$ and $Fn^{20}$ for *D=50* are mentioned in fig.2.

**Table 5.** Comparative analysis of various optimization algorithms with respect to mean *(μ)* standard deviation *(σ)*

| Test functions | | FF | SCA | PSO | ABC | CSCF |
|---|---|---|---|---|---|---|
| $Fn^1$ | *μ* | 5.82 E+08 | 5.68 E+05 | 7.21 E+02 | 2.57 E+02 | 1.56 E+02 |
| | *σ* | 2.40 E+07 | 7.11E+02 | 4.32 E+03 | 1.23 E+06 | 1.78 E+05 |
| $Fn^2$ | *μ* | 1.38 E+05 | 4.56 E+08 | 7.22 E+02 | 5.33 E+02 | 3.59E+03 |
| | *σ* | 4.21 E+04 | 6.27 E+02 | 4.45 E+08 | 2.45 E+03 | 5.49 E+03 |
| $Fn^3$ | *μ* | 2.24 E+03 | 7.46 E+02 | 7.31 E+02 | 6.38 E+02 | 3.65 E+04 |
| | *σ* | 3.56E+04 | 5.30 E+02 | 2.38 E+05 | 3.49 E+02 | 3.28 E+07 |
| $Fn^4$ | *μ* | 8.37 E+04 | 2.47 E+02 | 2.14 E+04 | 3.34 E+07 | 3.46 E+02 |
| | *σ* | 1.06 E+06 | 1.56 E+06 | 4.58 E+03 | 7.21 E+02 | 3.65 E+02 |
| $Fn^5$ | *μ* | 1.45 E+06 | 3.20E+07 | 7.06 E+02 | 5.27 E+03 | 3.86 E+07 |
| | *σ* | 3.04E+07 | 1.38 E+04 | 6.58 E+05 | 4.64 E+02 | 6.38E+05 |
| $Fn^6$ | *μ* | 4.28 E+03 | 2.54 E+07 | 3.91E+07 | 3.59 E+05 | 4.50E+03 |
| | *σ* | 3.71 E+01 | 7.45 E+02 | 3.28 E+04 | 4.37 E+07 | 3.28 E+03 |
| $Fn^7$ | *μ* | 1.37 E+04 | 7.38 E+07 | 1.49 E+05 | 3.54 E+10 | 7.47 E+06 |
| | *σ* | 6.39 E+03 | 3.71 E+07 | 2.83 E+04 | 2.19 E+07 | 1.49 E+04 |
| $Fn^8$ | *μ* | 1.34 E+02 | 2.13 E+05 | 1.48 E+04 | 3.41 E+07 | 2.49E+04 |
| | *σ* | 4.57 E+04 | 2.72E+02 | 3.51 E+07 | 7.35 E+05 | 2.46 E+03 |
| $Fn^9$ | *μ* | 3.27 E+06 | 6.46 E+02 | 6.72 E+07 | 4.24 E+02 | 2.62 E+03 |
| | *σ* | 3.67 E+03 | 4.39 E+06 | 1.27 E+06 | 7.48 E+04 | 3.25 E+03 |
| $Fn^{10}$ | *μ* | 2.17 E+05 | 2.42 E+02 | 2.50 E+04 | 2.48 E+02 | 2.69 E+05 |
| | *σ* | 4.57 E+03 | 4.04 E+02 | 2.37 E+07 | 4.76 E+02 | 2.59 E+05 |
| $Fn^{11}$ | *μ* | 9.52E+02 | 5.37 E+05 | 4.01 E+02 | 1.27 E+05 | 1.69 E+02 |
| | *σ* | 7.39 E+05 | 3.27 E+06 | 3.47 E+07 | 1.25 E+03 | 1.50 E+02 |
| $Fn^{12}$ | *μ* | 2.59 E+03 | 2.40 E+05 | 5.30 E+05 | 1.38 E+02 | 2.47 E+02 |
| | *σ* | 1.36 E+07 | 4.29 E+02 | 4.39 E+03 | 2.47 E+02 | 2.59 E+02 |
| $Fn^{13}$ | *μ* | 2.49 E+06 | 2.49 E+02 | 2.19 E+03 | 3.33E+07 | 3.48 E+05 |
| | *σ* | 3.50 E+02 | 2.28 E+03 | 5.27 E+01 | 3.49 E+06 | 1.39 E+09 |
| $Fn^{14}$ | *μ* | 2.28 E+06 | 4.48 E+03 | 1.59 E+06 | 3.27 E+09 | 1.67 E+05 |
| | *σ* | 2.46 E+09 | 7.95 E+23 | 3.29 E+04 | 5.39 E+04 | 5.28 E+02 |
| $Fn^{15}$ | *μ* | 1.39 E+06 | 3.31 E+02 | 4.87 E+02 | 2.49 E+08 | 4.68 E+02 |
| | *σ* | 1.47 E+04 | 6.38 E+08 | 5.82 E+08 | 2.48 E+06 | 2.47 E+10 |
| $Fn^{16}$ | *μ* | 4.61 E+02 | 4.33 E+02 | 6.37 E+02 | 7.37 E+02 | 2.15 E+04 |
| | *σ* | 2.37 E+06 | 4.54 E+02 | 1.23E+02 | 2.49 E+07 | 1.48 E+03 |
| $Fn^{17}$ | *μ* | 4.21 E+02 | 3.48 E+05 | 2.49 E+05 | 1.48 E+06 | 3.27E+03 |
| | *σ* | 4.28 E+08 | 2.50 E+04 | 1.23E+02 | 1.25 E+02 | 4.56 E+08 |
| $Fn^{18}$ | *μ* | 5.23 E+05 | 3.18 E+02 | 1.25 E+02 | 1.58 E+02 | 5.39 E+02 |
| | *σ* | 2.68 E+07 | 2.59 E+07 | 1.38 E+04 | 2.48 E+04 | 1.37 E+03 |
| $Fn^{19}$ | *μ* | 1.38 E+07 | 3.46 E+03 | 2.12 E+04 | 7.38 E+07 | 1.69 E+02 |
| | *σ* | 2.59 E+07 | 1.46 E+09 | 2.59 E+02 | 2.58 E+03 | 1.50 E+02 |
| $Fn^{20}$ | *μ* | 6.38 E+07 | 2.27 E+07 | 1.22 E+11 | 4.58 E+04 | 1.37 E+07 |
| | *σ* | 4.68 E+02 | 1.46 E+03 | 3.78 E+05 | 1.30 E+02 | 2.48 E+05 |

**Table 6.** Comparative analysis of various optimization algorithms with respect to Best value *(B)* and worst value *(W)*

| Test functions | | FF | SCA | PSO | ABC | CSCF |
|---|---|---|---|---|---|---|
| Fn[1] | *B* | 6.12E+05 | 2.09 E+08 | 5.64 E+08 | 4.51 E+08 | 1.38 E+04 |
| | *W* | 4.10 E+08 | 2.48 E+02 | 4.13 E+08 | 3.33 E+02 | 7.38 E+07 |
| Fn[2] | *B* | 1.32E+02 | 1.11E+05 | 2.59 E+04 | 7.12E+05 | 1.27 E+06 |
| | *W* | 1.59 E+06 | 3.27E+03 | 6.77E+05 | 3.34E+05 | 6.37 E+02 |
| Fn[3] | *B* | 7.37 E+02 | 2.17 E+05 | 3.34 E+07 | 1.78 E+02 | 2.83 E+04 |
| | *W* | 2.32 E+08 | 2.22 E+07 | 2.49 E+05 | 6.07E+05 | 1.48 E+06 |
| Fn[4] | *B* | 2.48 E+04 | 2.48 E+06 | 2.89 E+08 | 6.46 E+02 | 2.38 E+05 |
| | *W* | 7.13E+04 | 3.78 E+05 | 6.39 E+03 | 2.47 E+02 | 5.37 E+05 |
| Fn[5] | *B* | 5.49 E+03 | 1.37 E+03 | 4.13 E+08 | 2.19 E+03 | 1.37 E+04 |
| | *W* | 2.49 E+06 | 1.38 E+05 | 2.49E+04 | 7.47 E+06 | 1.27 E+06 |
| Fn[6] | *B* | 7.58 E+04 | 1.48 E+04 | 1.27 E+05 | 3.48 E+05 | 1.49 E+05 |
| | *W* | 1.37 E+07 | 1.25 E+02 | 3.29 E+04 | 7.11 E+04 | 3.48 E+05 |
| Fn[7] | *B* | 5.27 E+01 | 3.49 E+06 | 6.23E+05 | 4.45 E+08 | 2.59 E+07 |
| | *W* | 1.32 E+05 | 1.46 E+03 | 4.61 E+02 | 2.24 E+03 | 7.88 E+04 |
| Fn[8] | *B* | 7.38 E+07 | 2.69 E+05 | 6.58 E+05 | 5.68 E+05 | 2.46 E+09 |
| | *W* | 3.27 E+06 | 1.78 E+05 | 8.37 E+04 | 6.27 E+02 | 2.19 E+07 |
| Fn[9] | *B* | 1.23E+05 | 6.38 E+07 | 4.32 E+08 | 7.48 E+04 | 2.49 E+07 |
| | *W* | 1.34 E+02 | 2.51E+04 | 4.57 E+03 | 4.18 E+08 | 4.28 E+03 |
| Fn[10] | *B* | 4.20 E+08 | 2.08 E+02 | 1.51E+05 | 2.68 E+07 | 4.58 E+03 |
| | *W* | 1.58 E+02 | 1.22 E+11 | 2.08 E+07 | 1.38 E+07 | 2.27 E+07 |
| Fn[11] | *B* | 7.52 E+04 | 6.18E+05 | 5.30 E+02 | 2.59 E+07 | 2.59 E+02 |
| | *W* | 3.46 E+02 | 2.58 E+03 | 4.48 E+03 | 7.12 E+04 | 4.87 E+02 |
| Fn[12] | *B* | 3.59E+03 | 5.28 E+02 | 3.47 E+07 | 5.39 E+02 | 1.22 E+11 |
| | *W* | 1.56 E+02 | 1.67 E+05 | 2.12 E+04 | 3.28 E+03 | 6.38 E+08 |
| Fn[13] | *B* | 1.48 E+03 | 2.59 E+05 | 6.38 E+02 | 9.52E+02 | 1.56 E+06 |
| | *W* | 2.48 E+05 | 1.69 E+02 | 4.32 E+03 | 3.46 E+03 | 1.46 E+09 |
| Fn[14] | *B* | 1.50 E+02 | 2.02 E+08 | 4.57 E+04 | 1.55 E+05 | 2.50 E+04 |
| | *W* | 1.39 E+06 | 4.68 E+02 | 3.27 E+06 | 2.59 E+03 | 3.67 E+03 |
| Fn[15] | *B* | 2.28 E+06 | 1.30 E+02 | 1.23E+02 | 3.50 E+02 | 7.95 E+23 |
| | *W* | 5.39 E+04 | 3.28 E+04 | 2.46 E+03 | 2.99 E+08 | 4.28 E+08 |
| Fn[16] | *B* | 6.33E+05 | 2.11 E+08 | 2.15 E+04 | 1.49 E+04 | 2.50 E+04 |
| | *W* | 4.58 E+04 | 2.45 E+03 | 1.01 E+05 | 6.20E+05 | 3.49 E+02 |
| Fn[17] | *B* | 1.36 E+07 | 1.38 E+02 | 1.23 E+06 | 2.47 E+02 | 3.31 E+02 |
| | *W* | 4.56 E+08 | 4.39 E+03 | 2.62 E+03 | 1.20 E+05 | 4.39 E+06 |
| Fn[18] | *B* | 1.36 E+06 | 2.49 E+08 | 3.18 E+02 | 2.37 E+06 | 3.54 E+10 |
| | *W* | 2.47 E+02 | 2.59 E+02 | 6.38E+05 | 5.33 E+02 | 2.28 E+03 |
| Fn[19] | *B* | 4.56 E+08 | 4.37 E+07 | 4.50E+03 | 2.49 E+02 | 2.54 E+07 |
| | *W* | 7.39 E+05 | 3.28 E+07 | 4.68 E+02 | 2.13 E+05 | 2.11 E+08 |
| Fn[20] | *B* | 5.23 E+05 | 2.47 E+10 | 3.86 E+07 | 1.39 E+09 | 5.30 E+05 |
| | *W* | 1.47 E+04 | 2.40 E+05 | 2.37 E+07 | 3.27 E+09 | 5.27 E+03 |

### *4.3 Comparison of CSCF with various optimization algorithms*

This section demonstrates the comparative analysis of the proposed CSCF algorithm with various other optimization algorithms such as FireFly (FF) algorithm [19], Sine Cosine Algorithm (SCA) [20], Particle Swarm Optimization (PSO) approach [23], Artificial Bee Colony (ABC) optimization algorithms [24] to evaluate the effectiveness of the CSCF algorithm. Here, the dimension value range is assumed to be *D=100*; where each algorithm is implemented by using 20 benchmark functions. Moreover, the size of the initial population is set as 20. Therefore, the comparative analysis of various optimization approaches with respect to mean *(μ)* standard deviation *(σ)* is explained in Table 5. Also, the comparative analysis for various optimization approaches with respect to Best value *(B)* and worst value *(W)* is explained in Table 6. The experimental analysis based on average rank testing of the CSCF approach is compared with FF, SCA, PSO and ABC in Table 7. Therefore, the experimental results for two different types of tests namely Wilcoxon's rank-sum (R-S) test and Wilcoxon's multiple problem (M-P) tests for the proposed CSCF algorithm based on twenty benchmark functions are described in Table 7. In addition to this, the value of $\mathrm{P} < (\varepsilon = 0.1 \, and \, 0.05)$ and $r^{+} \, and \, r^{-}$ values are greater for all respective cases. Thus the analysis reveals that the proposed CSCF algorithm provides better performances when compared to all other approaches [26].

**Table 7.** Analysis of Wilcoxon's (R-S) test and Wilcoxon's (M-P) tests of CSCF algorithm

| Approaches | Best Value (B) | Worst value (W) | $r^{+}$ | $r^{-}$ | P | €=0.1 | €=0.05 |
|---|---|---|---|---|---|---|---|
| CSCF $v_S$ FF | 16 | 4 | 201 | 41 | 2.46E-02 | YES | YES |
| CSCF $v_S$ SCA | 12 | 8 | 104 | 24 | 5.23E-03 | YES | YES |
| CSCF $v_S$ PSO | 19 | 1 | 211 | 7 | 7.75E-02 | YES | YES |
| CSCF $v_S$ ABC | 11 | 9 | 223 | 83 | 3.61E-01 | YES | YES |

### *4.4 Real-time engineering design problem for CSCF*

In this section, the efficiency and the performances of the proposed CSCF algorithm is solved by evaluating three different types of engineering design problems namely Welded Beam Design $(WB_D)$, Pressure Vessel Design $(PV_D)$ and Tension-Compression Spring Design $(T - CS_D)$. These engineering design problems are described briefly in the following section. Here, the initial size of the population is set as 20 and the maximum size of the population is set as 100.

## A. *Illustration 1: Problem based on Welded Beam Design* $(WB_D)$

This section illustrates problem based on the design of a welding beam $(WB_D)$ [27]; where the minimum cost function is subjected to several constraints namely the beam's end deflection $(BE_D)$, beam's bending stress $(BB_S)$, Shear stress $(S_S)$ and buckling load of the bar $(BB_L)$. Moreover, the $(WB_D)$ comprises of four different types of variables namely $H(Z_1)$, $L(Z_2)$, $T(Z_3)$ and $B(Z_1)$ respectively. The structural model for the problem based on a welding beam $(WB_D)$ is represented in fig.3.

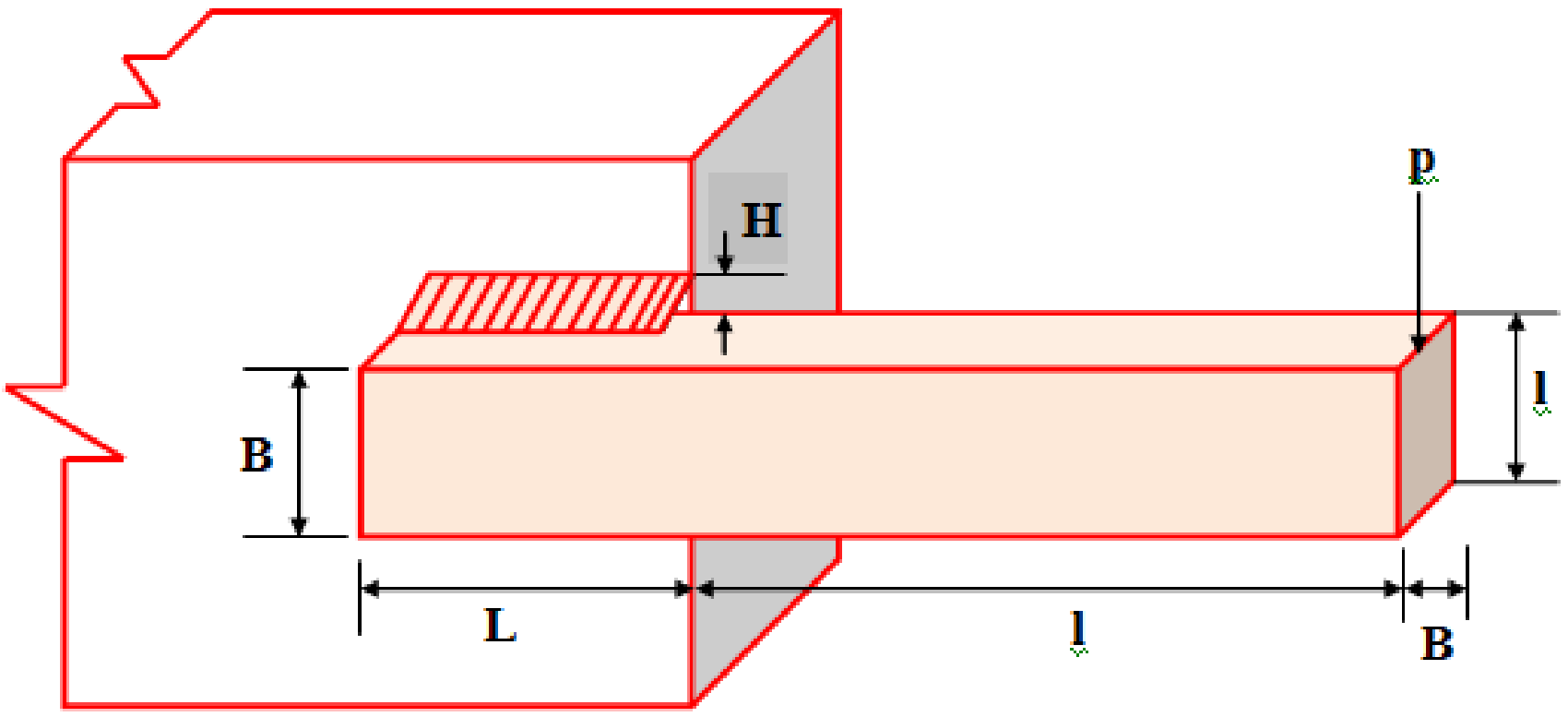

**Fig.3** Design for the welding beam $(WB_D)$ problem

The mathematical expression based on the design for the problem based on a welding beam $(WB_D)$ is formulated in the following section.

$$Minimize: \ F(Z) = 1.1047 L Z_1^2 Z_2 + 0.04811 Z_3 Z_4 (14.0 + Z_2) \tag{21}$$

*Subject to*:

$$G_1(Z) = S_S(Z) - S_S^{MAX} \leq 0 \tag{22}$$

$$G_2(Z) = BB_S(Z) - BB_S^{MAX} \leq 0 \tag{23}$$

$$G_3(Z) = Z - 1 - Z_4 \leq 0 \tag{24}$$

$$G_4(Z) = 1.1047 L Z_1^2 + 0.04811 Z_3 Z_4 (14.0 + Z_2) - 5.0 \leq 0 \tag{25}$$

$$G_5(Z) = 0.125 - Z_1 \leq 0 \tag{26}$$

$$G_6(Z) = BE_D(Z) - BE_D^{MAX} \leq 0 \tag{27}$$

$$G_7(Z) = p - BB_L(Z) \leq 0 \tag{28}$$

Therefore, the expression for several constraints and variables based on the problem based on the design of a welding beam $(WB_D)$ are delineated in the following section.

$$S_S(Z) = \sqrt{\left(S_S^{'}\right)^2 + 2S_S^{'}S_S^{''}\frac{Z_2}{2r} + \left(S_S^{''}\right)^2} \tag{29}$$

From the above equation,

$$S_S^{'} = \frac{p}{\sqrt{2}\,Z_1Z_2};\ \ S_S^{''} = \frac{Jr}{M};\ \ J = P\left(l + \frac{Z_2}{2}\right);\ BB_S(Z) = \frac{6pl}{Z_4\,Z_3^2}$$

$$BE_D(Z) = \frac{4pl^3}{eZ_4\,Z_3^3};\ \ M = 2\sqrt{2}\,Z_1Z_2\left[\frac{Z_2^2}{12} + \left(\frac{Z_1+Z_3}{2}\right)^2\right]$$

$$r = \sqrt{\frac{Z_2^2}{4} + \left(\frac{Z_1+Z_2}{2}\right)^2};\ \ BB_L(Z) = \frac{4.013e\sqrt{\frac{Z_3^2Z_4^6}{36}}}{l^2}\left(1 - \frac{Z_3}{2l}\sqrt{\frac{e}{4F}}\right)$$

$$p = 6000LB;\ l = 14in;\ e = 30\times10^6\ psi;\ F = 12\times10^6\ psi$$

$$S_S^{MAX} = 1360\ psi \quad BB_S^{MAX} = 3000\ psi \quad BE_D^{MAX} = 0.25psi$$

Table 8 provides the best solutions for various approaches such as FF, SCA, PSO, ABC and proposed CSCF approaches. Table 9 provides the statistical analysis for the mean *(μ)* standard deviation *(σ)*, Best value *(B)* and worst value *(W)*.

**Table 8.** Best solutions for various approaches based on $WB_D$

| Variables | CSCF | FF | SCA | PSO | ABC |
|---|---|---|---|---|---|
| F(Z) | 1.704 | 2.236 | 1.725 | 1.942 | 2.358 |
| $Z_1$(H) | 0.197 | 1.237 | 1.478 | 2.365 | 0.937 |
| $Z_2$(L) | 8.035 | 7.238 | 5.323 | 8.368 | 7.234 |
| $Z_3$(T) | 3.209 | 4.235 | 3.736 | 3.897 | 5.237 |
| $Z_4$(B) | 2.210 | 4.358 | 2.789 | 3.247 | 4.374 |
| $G_1$(Z) | -4.288 | -5.235 | -5.565 | -7.327 | -5.856 |
| $G_2$(Z) | -4.789 | -5.385 | -6.462 | NA | -7.357 |
| $G_3$(Z) | -0.499 | -1.375 | -1.458 | NA | -2.345 |
| $G_4$(Z) | -0.067 | -0.475 | -0.637 | NA | -0.927 |
| $G_5$(Z) | -3.274 | -3.985 | -4.214 | NA | -4.274 |
| $G_6$(Z) | -3.173 | -4.763 | -5.247 | NA | -6.436 |
| $G_7$(Z) | -2.438 | -3.475 | -2.462 | -4.287 | -3.345 |

**Table 9.** Statistical analysis of various approaches for $WB_D$

| Approaches | $\mu$ | $\sigma$ | $B$ | $W$ |
|---|---|---|---|---|
| CSCF | 1.7043 | 1.7042 | 1.7048 | 1.7044 |
| FF | NA | NA | 2.236432 | NA |
| SCA | 1.7256 | 1.72564 | 1.7258 | 1.7252 |
| PSO | NA | NA | 1.942762 | NA |
| ABC | 2.3585 | 2.3583 | 2.3584 | 1.7487 |

*B. Illustration 2: Problem based on Pressure Vessel Design* $(PV_D)$

The problem based on Pressure Vessel Design $(PV_D)$ aims in minimizing the manufacturing cost function [28]. The structural design of the pressure vessel is represented in fig. 4 that contains the working pressure and the volume of about 3000 psi and 750 ft$^3$. In addition to this, the Pressure Vessel Design $(PV_D)$ comprises of four different variables namely the shell thickness $S_T$ as $Z_1$, Head thickness $H_T$ as $Z_2$, Inner radius $I_R$ as $Z_3$, the cylindrical section having the length $l$ as $Z_4$. Here, the continuous variables are denoted as $Z_3$ and $Z_4$ where the integral multiples are denoted as $Z_1$ and $Z_2$ respectively. Then the mathematical expression based on the design for the problem based on Pressure Vessel Design $(PV_D)$ is formulated in the following section.

$$Minimize: \ F(Z) = 0.6224 Z_1 Z_3 Z_4 + 1.7781 Z_2 Z_3^2 + 3.1611 Z_1^2 Z_4 + 19.84 Z_1^2 Z_3 \quad (30)$$

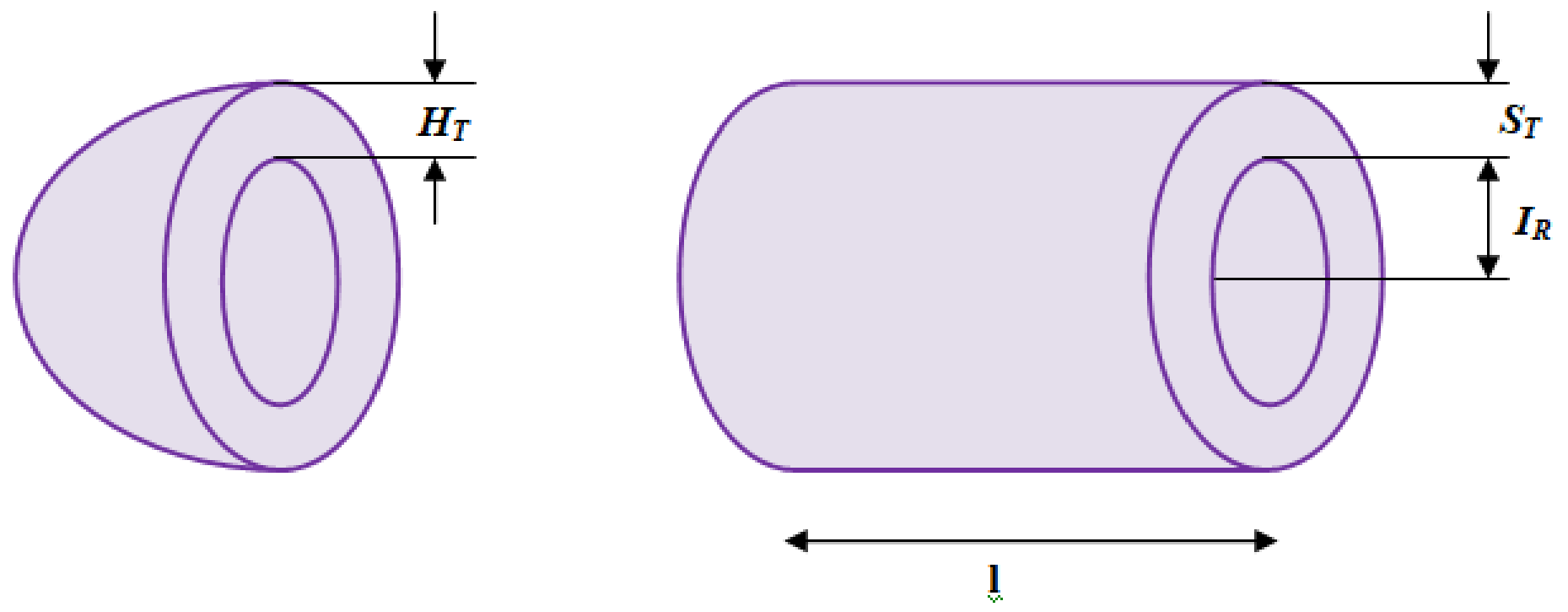

**Fig.4** Design for Pressure Vessel Design $(PV_D)$ problem

*Subject to*:

$$G_1(Z_1, Z_3) = -Z_1 + 0.0193 Z_3 \le 0 \quad (31)$$

$$G_2(Z_2, Z_3) = -Z_2 + 0.0095 Z_3 \le 0 \quad (32)$$

$$G_3(Z_3, Z_4) = -\pi Z_3^2 Z_4 - \frac{4}{3}\pi Z_3^2 + 1296000 \le 0 \quad (33)$$

$$G_4(Z_4) = Z_4 + 240 \le 0 \quad (34)$$

Table 10 provides the best solutions for various approaches such as FF, SCA, PSO, ABC and proposed CSCF approaches. Table 11 provides the statistical analysis for the mean *(μ)* standard deviation *(σ)*, Best value *(B)* and worst value *(W)*.

**Table 10.** Best solutions for various approaches based on $PV_D$

| Variables | CSCF | FF | SCA | PSO | ABC |
|---|---|---|---|---|---|
| F(Z) | 6123.489 | 6356.956 | 647.025 | 6485.382 | 6627.827 |
| $Z_1(S_T)$ | 0.726329 | 0.729647 | 0.74583 | 0.75938 | 0.763045 |
| $Z_2(H_T)$ | 0.527452 | 0.537219 | 0.53947 | 0.54728 | 0.54682 |
| $Z_3(I_R)$ | 41.66390 | 41.86719 | 42.4893 | 43.4710 | 44.3729 |
| $Z_4(l)$ | 163.4489 | 163.5762 | 164.294 | 165.328 | 166.320 |
| $G_1$(Z) | -0.000147 | -0.000163 | 0.000245 | NA | 0.00038 |
| $G_2$(Z) | -0.043820 | -0.044625 | 0.043894 | NA | 0.04429 |
| $G_3$(Z) | -112.4896 | -112.5782 | 113.4340 | NA | 114.927 |
| $G_4$(Z) | -60.47343 | -60.62632 | -61.3829 | NA | -62.3840 |

**Table 11.** Statistical analysis of various approaches for $PV_D$

| Approaches | *μ* | *σ* | *B* | *W* |
|---|---|---|---|---|
| CSCF | 6123.489 | 2.427594 | 6123.532 | 6123.563 |
| FF | 6034.87 | 83.26723 | 635.809 | 6234.87 |
| SCA | NA | NA | 622.479 | NA |
| PSO | NA | NA | 632.479 | NA |
| ABC | 6142.763 | 12.3769 | 6782.498 | 6232.457 |

*C. Illustration 3: Problem based on Tension-Compression Spring Design* $(T - CS_D)$

Fig.5 describes the structural model for the problem based on Tension-Compression Spring Design $(T - CS_D)$. Here, the $(T - CS_D)$ is considered as one of the continuous constrained problem developed by Belegundu [29]. Moreover, the Tension-Compression Spring Design $(T - CS_D)$ comprises of four different parameters namely diameter of the coil $(D_C)$, active coil number $(N_C)$ with the diameter $(D)$. Then the mathematical expression based on the design for the problem based on Tension-Compression Spring Design $(T-CS_D)$ is formulated in the following section.

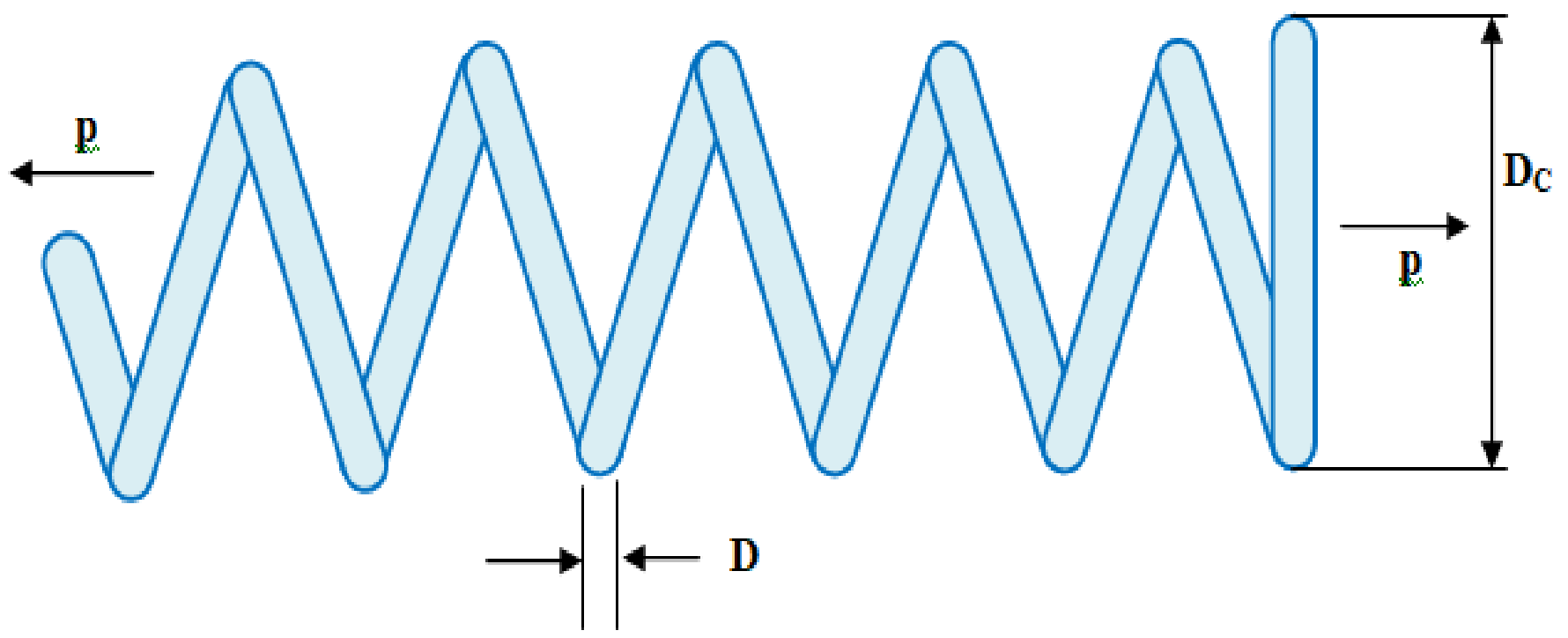

**Fig.5** Design for Tension-Compression Spring Design $(T-CS_D)$ problem

$$\text{Let us assume, } Z = [Z_1, Z_2, Z_3] = D_C, N_C, D \tag{35}$$

$$Minimize: \; F(Z) = [N_C + 2]D_C D^2 \tag{36}$$

*Subject to*:

$$G_1(Z) = 1 - \frac{D_C^3 N_C}{71785D} \leq 0 \tag{37}$$

$$G_2(Z) = \frac{4D_C^2 - DD_C}{12566(D_C D^2 - D^4)} + \frac{1}{5108D^2} - 1 \leq 0 \tag{38}$$

$$G_3(Z) = 1 - \frac{140.45D}{N_C D^2} \leq 0 \tag{39}$$

$$G_4(Z) = \frac{D + D_C}{1.5} - 1 \leq 0 \tag{40}$$

Then the design variables for the problem based on Tension-Compression Spring Design $(T-CS_D)$ are delineated in the following section. $0.05 \leq D \leq 2$; $0.25 \leq D_C \leq 1.3$ as well as $2 \leq D_C \leq 15$. Table 12 provides the best solutions for various approaches such as FF, SCA, PSO, ABC and proposed CSCF approaches. Table 13 provides the statistical analysis for the mean *(μ)* standard deviation *(σ)*, Best value *(B)* and worst value *(W)*.

**Table 12`.** Best solutions for various approaches based on $T-CS_D$

| Variables | CSCF | FF | SCA | PSO | ABC |
|---|---|---|---|---|---|
| F(Z) | 0.020342 | 0.028833 | 0.022856 | 0.0265978 | 0.027573 |
| $Z_1(D_C)$ | 0.374584 | 0.374637 | 0.383674 | 0.3936732 | 0.426537 |
| $Z_2(N_C)$ | 0.503762 | 0.527482 | 0.53842 | 0.543785 | 0.568239 |
| $Z_3(D)$ | 10.83740 | 10.8643 | 11.0352 | 11.36789 | 11.48793 |
| $Z_4$(B) | -5.37998 | -5.43789 | -5.38265 | -5.401632 | -5.41789 |
| $G_1$(Z) | -3.89787 | -4.67882 | -4.7298 | NA | -4.28701 |
| $G_2$(Z) | -0.26379 | -0.27245 | -0.28753 | NA | -0.29363 |
| $G_3$(Z) | -4.67903 | -4.83567 | -4.69365 | NA | -4.72734 |
| $G_4$(Z) | -0.76727 | -0.77346 | -0.78437 | NA | -0.72763 |

**Table 13.** Statistical analysis of various approaches for $T-CS_D$

| Approaches | $\mu$ | $\sigma$ | $B$ | $W$ |
|---|---|---|---|---|
| CSCF | 0.021356 | 3.67811 | 0.021356 | 0.021356 |
| FF | 0.021356 | 4.3689 | 0.021356 | 0.021356 |
| SCA | NA | NA | 0.021356 | NA |
| PSO | NA | NA | 0.021356 | NA |
| ABC | 0.021356 | 4.58678 | 0.021356 | 0.021356 |

Then, various chaotic variants namely

Variant- I, Variant- II, Variant- III, Variant- IV, and Variant- V of the novel CSCF algorithm is employed in solving the above mentioned three engineering problems namely $P_1, P_2, P_3$ that are represented in Table 14. Fig.6 explains the ranking system of five different variants of the CSCSF algorithm. From the graphical analysis, it is noted that the fourth variant of the circle mapping provides minimum Mean Absolute Error (MAE) when compared with all other variants. The term MAE is defined as the average value obtained by the absolute difference among the actual value and the predicted value. Finally, fig.7 describes the convergence time with respect to all the five variants. The graphical analysis reveals that the variant-II comprises of less convergence time during the running process when compared with all other variants.

**Table 14.** CM optimization for six variants of CSCF algorithm

| Problems | CM | Variant I | Variant II | Variant III | Variant IV | Variant V |
|---|---|---|---|---|---|---|
| $P_1$ | Logistic | 0.23024 | 0.23453 | 0.25687 | 0.22475 | 0.22946 |
| | Tent | 0.26437 | 0.26892 | 0.27356 | 0.23728 | 0.26409 |
| | Sinusoidal | 0.29037 | 0.29784 | 0.29487 | 0.25782 | 0.27365 |
| | Gauss | 0.24893 | 0.24632 | 0.24387 | 0.24023 | 0.24973 |
| | Circle | 0.22731 | 0.22472 | 0.22636 | **0.22022** | 0.22537 |
| | Sinus | 0.24577 | 0.24376 | 0.24937 | 0.24065 | 0.24637 |
| | Iterative | 0.23263 | 0.23436 | 0.24854 | 0.23036 | 0.24872 |
| | Chebyshev | 0.26972 | 0.26253 | 0.27463 | 0.26165 | 0.27261 |
| | Henon | 0.23365 | 0.23343 | 0.24876 | 0.23347 | 0.23226 |
| | Intermittency | 0.25376 | 0.25434 | 0.26362 | 0.25336 | 0.26328 |
| | Singer | 0.26287 | 0.25421 | 0.27887 | 0.26114 | 0.27115 |
| | Sine | 0.24763 | 0.24376 | 0.24874 | 0.24462 | 0.24736 |
| $P_2$ | Logistic | 0.33536 | 0.33543 | 0.33472 | 0.33398 | 0.33253 |
| | Tent | 0.33374 | 0.33346 | 0.33464 | 0.33487 | 0.33376 |
| | Sinusoidal | 0.34476 | 0.34236 | 0.34345 | 0.34212 | 0.34864 |
| | Gauss | 0.33536 | 0.33453 | 0.33463 | 0.33534 | 0.33562 |
| | Circle | 0.33342 | 0.33367 | 0.33376 | **0.33333** | 0.33364 |
| | Sinus | 0.34764 | 0.34463 | 0.34475 | 0.34874 | 0.34497 |
| | Iterative | 0.34373 | 0.34473 | 0.34472 | 0.34364 | 0.34398 |
| | Chebyshev | 0.33236 | 0.33372 | 0.33447 | 0.33243 | 0.33348 |
| | Henon | 0.34747 | 0.34248 | 0.34834 | 0.34172 | 0.34152 |
| | Intermittency | 0.34362 | 0.34235 | 0.34236 | 0.34234 | 0.34086 |
| | Singer | 0.33283 | 0.33263 | 0.33476 | 0.33107 | 0.33127 |
| | Sine | 0.34873 | 0.34473 | 0.34331 | 0.34163 | 0.34836 |
| $P_3$ | Logistic | 0.52243 | 0.52454 | 0.52345 | 0.52032 | 0.52257 |
| | Tent | 0.53476 | 0.53365 | 0.53673 | 0.53163 | 0.53256 |
| | Sinusoidal | 0.52677 | 0.52256 | 0.52488 | 0.52112 | 0.52157 |
| | Gauss | 0.52143 | 0.52657 | 0.52225 | 0.52376 | 0.52478 |
| | Circle | 0.52589 | 0.52144 | 0.52586 | **0.52002** | 0.52164 |
| | Sinus | 0.54254 | 0.54685 | 0.54148 | 0.54265 | 0.54574 |
| | Iterative | 0.53547 | 0.53148 | 0.53658 | 0.53356 | 0.53467 |
| | Chebyshev | 0.53679 | 0.53251 | 0.53135 | 0.53673 | 0.53682 |
| | Henon | 0.52146 | 0.52367 | 0.52785 | 0.52576 | 0.52147 |
| | Intermittency | 0.54457 | 0.54237 | 0.54652 | 0.54134 | 0.54658 |
| | Singer | 0.52254 | 0.52475 | 0.52111 | 0.52457 | 0.52537 |
| | Sine | 0.53789 | 0.53645 | 0.53467 | 0.53362 | 0.53364 |

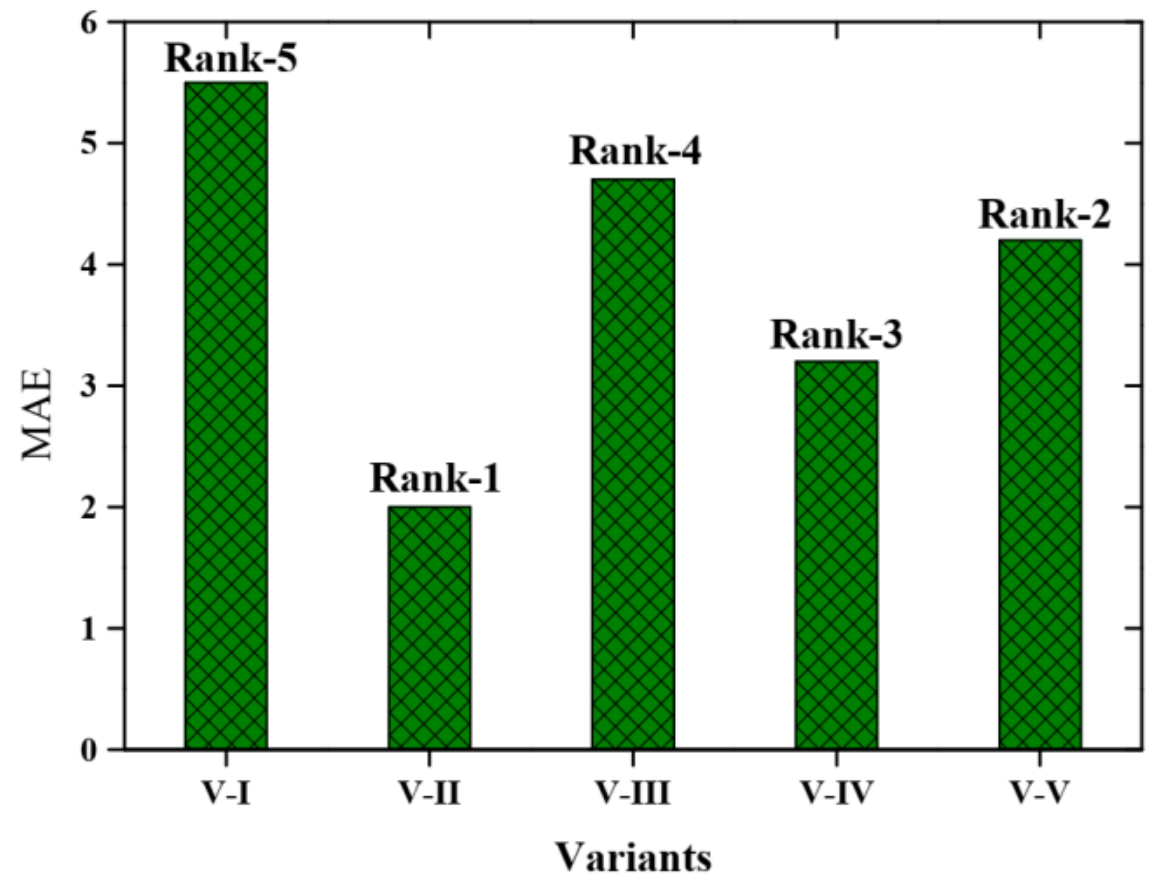

(a)

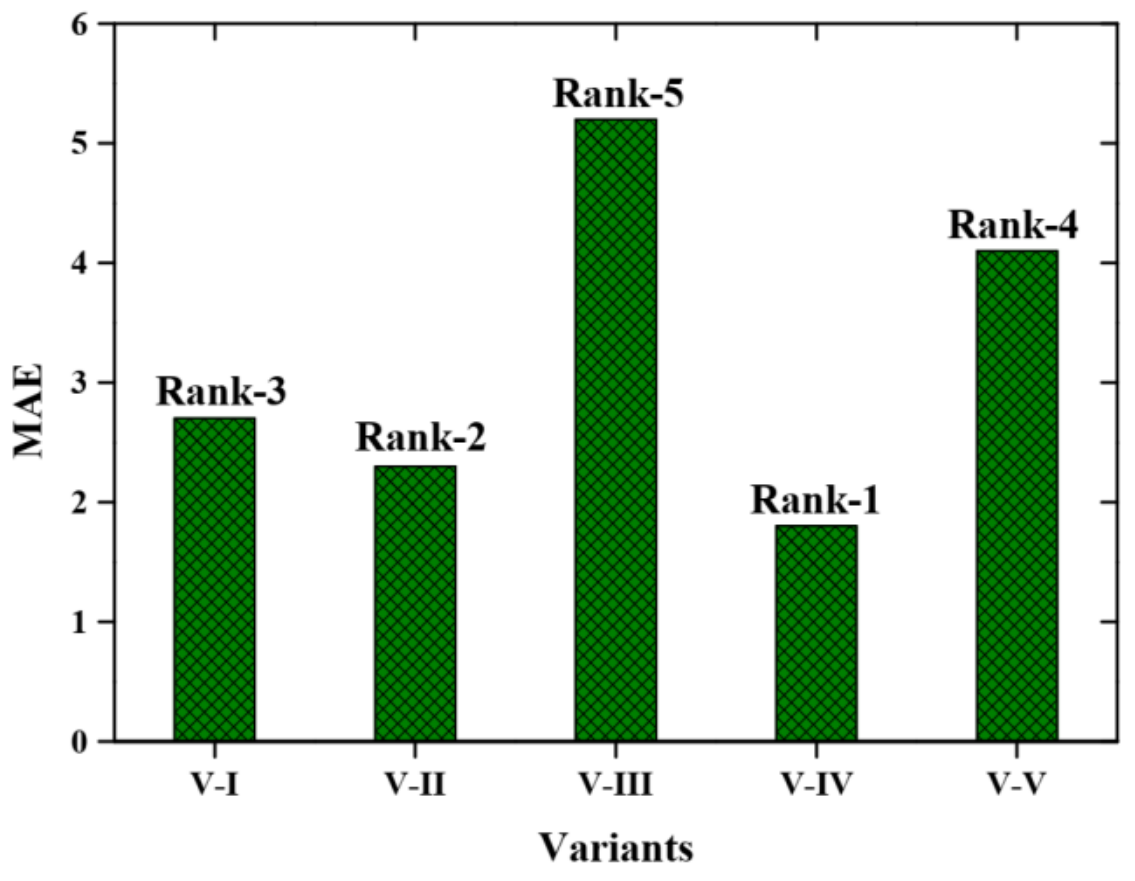

(b)

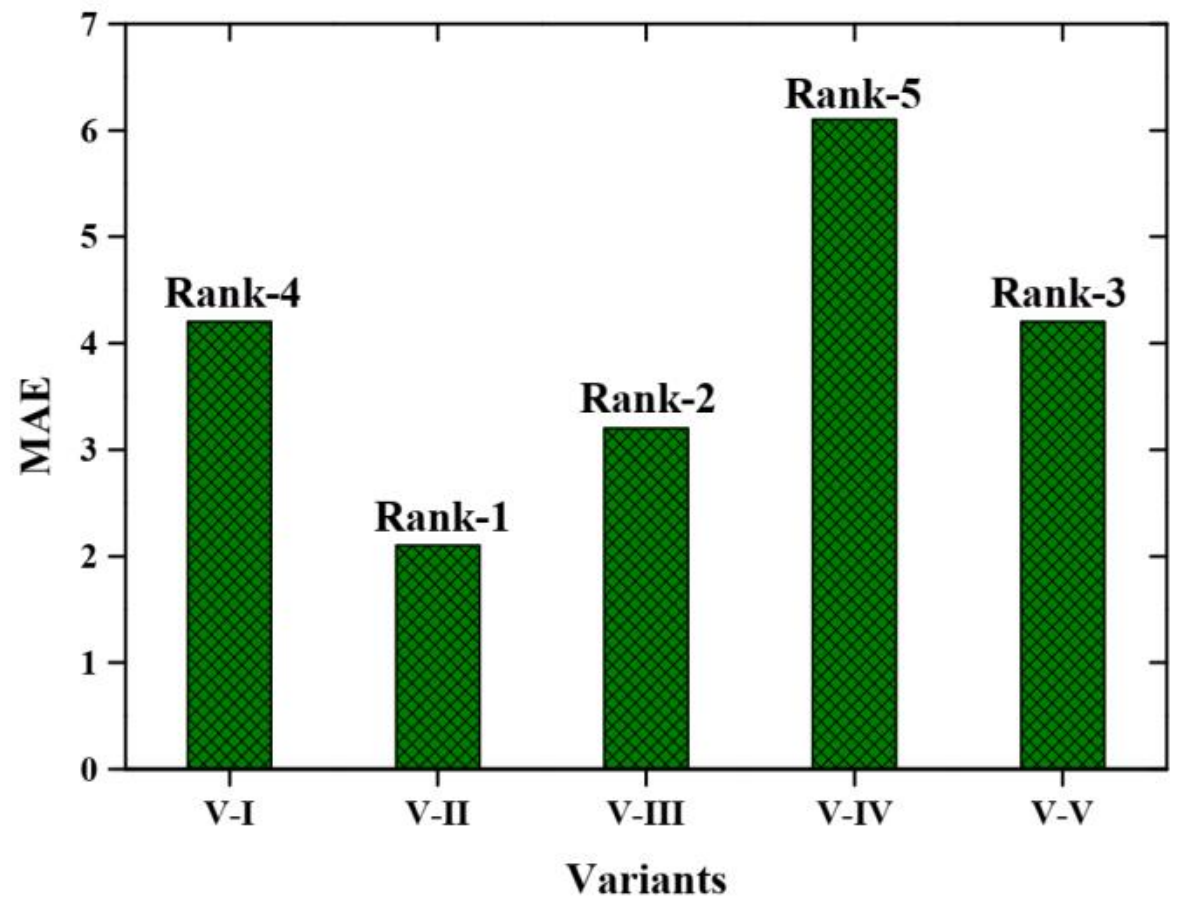

(c)

**Fig.6** MAE versus Variants for a) $P_1$ b) $P_2$ c) $P_3$

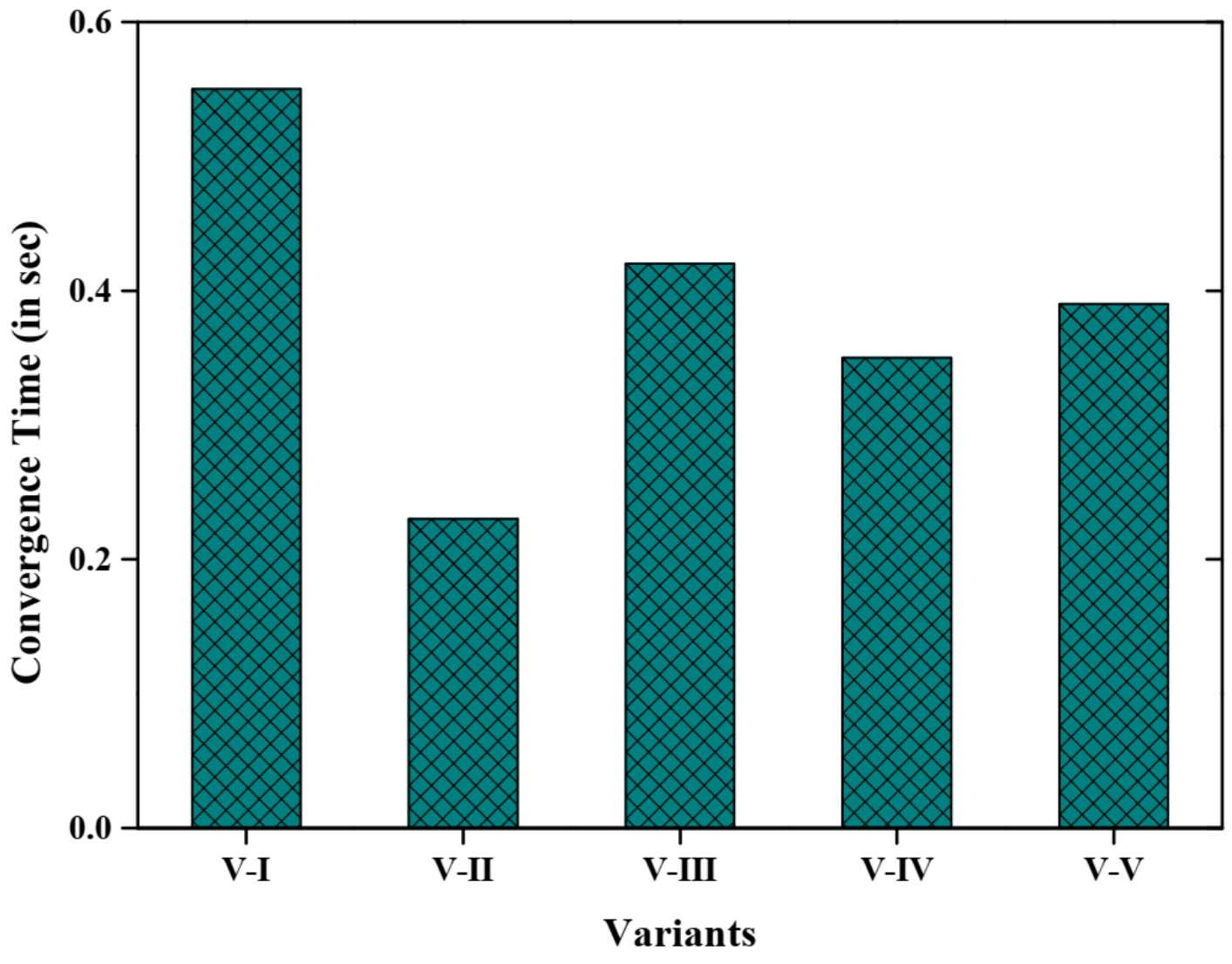

**Fig.7.** Convergence time versus variants

## 5. Conclusion

This paper proposed a novel Chaotic Sine Cosine Firefly (CSCF) algorithm with numerous variants to solve optimization problems such as computational complexity, memory space, tricky derivations, efficiency, convergence speed etc. The chaotic form of two algorithms namely the sine cosine algorithm (SCA) and the Firefly (FF) algorithms are integrated to improve the convergence speed and efficiency to minimize the complexity issues. Moreover, the proposed CSCF approach is operated under various chaotic phases and the optimal chaotic variants containing the best chaotic mapping is selected. Then various experiments are conducted to evaluate the efficiency and the performances of the Chaotic Sine Cosine Firefly (CSCF) algorithm. Owing to its hypothetical nature, various chaotic functions and benchmark functions are discussed to obtain better optimal results. Furthermore, the proposed CSCF algorithms are compared with several other optimization algorithms such as FireFly (FF) algorithm, Particle Swarm Optimization (PSO) approach, Artificial Bee Colony (ABC) optimization algorithms to evaluate the effectiveness of the CSCF algorithm. Finally, the efficiency and the performances of the proposed CSCF algorithm is solved by evaluating three different types of engineering design problems to prove the efficiency, robustness and effectiveness of the system.

## Compliance with ethical standards

### *Conflict of interest*

The authors declare that they have no conflict of interest.